\documentclass[journal]{IEEEtran}
\usepackage{times}
\usepackage{epsfig}
\usepackage{graphicx}
\usepackage{amsmath}
\usepackage{amssymb}
\usepackage{xspace}
\usepackage{subcaption}
\usepackage{multirow}
\usepackage{color}

\makeatletter
\DeclareRobustCommand\onedot{\futurelet\@let@token\@onedot}
\def\@onedot{\ifx\@let@token.\else.\null\fi\xspace}

\def\eg{\emph{e.g}\onedot} 
\def\ie{\emph{i.e}\onedot} 
 
\def\etc{\emph{etc}\onedot} 
 
\def\etal{\emph{et al}\onedot}
\makeatother

\newcommand{\ProblemAbbr}{MDE}

\newcommand{\MethodName}{Approach}

\newcommand{\hh}{\textcolor{black}}

\ifCLASSINFOpdf
\else
\fi
\begin{document}
%
\title{A Compromise Principle in Deep Monocular Depth Estimation}
%
%
%

\author{
Huan~Fu,
        Mingming~Gong,
        Chaohui~Wang,
        and~Dacheng~Tao,~\IEEEmembership{Fellow,~IEEE}
\thanks{H. Fu and D. Tao are with UBTECH Sydney AI Centre, SIT, FEIT, The University of Sydney,
J12 Cleveland St, Darlington NSW 2008, Australia (e-mail: hufu6371@uni.sydney.edu.au;
dacheng.tao@sydney.edu.au).}
\thanks{M. Gong is with 
Department of Biomedical Informatics, University of Pittsburgh, Cubicle 520c, 5607 Baum Bouevard, Pittsburgh, PA 15206 (e-mail:minggongnju@gmail.com).}
\thanks{C. Wang is with Laboratoire d'Informatique Gaspard Monge - CNRS
UMR 8049, Universit$\acute{\text{e}}$ Paris-Est, 77454 Marne-la-Vall$\acute{\text{e}}$e Cedex 2,
France (e-mail: chaohui.wang@u-pem.fr).}
}

\maketitle

\begin{abstract}

Monocular depth estimation, which plays a key role in understanding 3D scene geometry, is fundamentally an ill-posed problem. 
Existing methods based on deep convolutional neural networks (DCNNs) have examined this problem by learning convolutional networks to estimate continuous depth maps from monocular images. However, we find that training a network to predict a high spatial resolution continuous depth map often suffers from poor local solutions. In this paper, we hypothesize that achieving a compromise between spatial and depth resolutions can improve network training. Based on this ``compromise principle", we propose a regression-classification cascaded network (RCCN), which consists of a regression branch predicting a low spatial resolution continuous depth map and a classification branch predicting a high spatial resolution discrete depth map. The two branches form a cascaded structure allowing the main classification branch to benefit from the auxiliary regression branch. By leveraging large-scale raw training datasets and some data augmentation strategies, our network achieves competitive or state-of-the-art results on three challenging benchmarks, including NYU Depth V2 \cite{Silberman:ECCV12}, KITTI \cite{Geiger2013IJRR}, and Make3D \cite{saxena2009make3d}.
\end{abstract}


%
\IEEEpeerreviewmaketitle

\section{Introduction}
\label{sec:intro}
%
%
%
%
\IEEEPARstart{E}{stimating} depth from 2D images is a key component of scene reconstruction and understanding tasks, such as 3D recognition, tracking, segmentation and detection \cite{Sheng2017AGM, savarese20073d, sun2010layered, hoiem2005automatic, saxena2009make3d, sun2014rate, muller20133d, chung2014bit, hou2016deeply, kiechle2013joint}, \etc. In this paper, we examine the problem of~\emph{Monocular Depth Estimation}~(abbr. as \emph{\ProblemAbbr} hereafter), namely the estimation of the depth map from a single image.


Compared to depth estimation from stereo images or video sequences, in which significant progress has been made \cite{ha:cvpr16, Kong_2015_ICCV, liu2009continuous, poostchi2016semantic, chung2014bit, li2015depth_v2, karsch2014depth, rajagopalan2004depth},~progress in \ProblemAbbr~ has been slow. 
\ProblemAbbr~is fundamentally an ill-posed problem: a single 2D image may be produced from an infinite number of distinct 3D scenes. Fortunately, the 2D image and the depth map are correlated, suggesting that the depth can still be predicted with considerable accuracy.



To overcome this inherent ambiguity, typical methods resort to exploiting statistically meaningful monocular cues or features, such as perspective and texture information, object sizes, object locations, and occlusions. Previous methods used handcrafted features for depth estimation \cite{saxena2009make3d,hoiem2007recovering,ladicky2014pulling, konrad20122d, alatan1998estimation, lai2010metric, yang2013depth, tosic2014light, lin2013absolute, Dong2017ColorGuidedDR, Sheng2017GeometricOA}, but since the handcrafted features alone can only capture local information, probabilistic graphic models \cite{saxena2006learning, saxena2009make3d} or depth transfer methods  \cite{karsch2014depth} have been introduced to incorporate long range global cues.

Buoyed by the success of deep convolutional neural networks (DCNNs) in object recognition and detection, several recent works have significantly improved the~\ProblemAbbr~performance  by a large margin with the use of DCNN-based models \cite{liu2015deep, wang2015towards, roymonocular, eigen2015predicting, kim2016unified, kuznietsov2017semi}, demonstrating that deep features are superior to handcrafted features. The main advantage of DCNNs is that the hierarchical representations in a DCNN capture both local and global information. A state-of-the-art method \cite{eigen2015predicting} exploits the multi-scale network which firstly learns to predict a coarse depth map using global information and then refines it using another network with local information to produce a fine depth map.

Existing methods address the MDE problem by learning a CNN to estimate the continuous depth map. Since this problem is a standard regression problem, existing methods usually adopt the root mean squared error (RMSE) in log-space as the loss function. Although training with RMSE can achieve a reasonable solution when predicting a low resolution depth map, we find that the optimization tends to be difficult when we try to train networks to predict high-resolution continuous maps. The stochastic gradient descent (SGD) optimization method usually produces a local solution with unsatisfactory training error in this case. 

We hypothesize that a compromise between spatial and depth resolution can make the optimization easier, which is referred to as the ``compromise principle" in this paper. According to the compromise principle, we avoid directly estimating a high spatial resolution continuous depth map by firstly estimating depth maps with reduced spatial or depth resolution. To reduce the depth resolution, we propose to transform the regression problem into a classification problem by discretizing the depth value into intervals. Employing the classification loss to train the network achieves lower RMSE on the training data than training with RMSE. Also, the low spatial resolution continuous map can be learned with considerable accuracy. Based on such a principle, we develop a regression-classification cascade network (RCCN) which consists of two branches: 1) the regression branch predicting low spatial resolution continuous depth map from the fully-connected layers, capable of capturing global scene information; and 2) the classification branch, which predicts high spatial resolution discrete depth maps from the convolutional layers to preserve finer spatial information. The two branches form a cascaded structure and are learned jointly in an end-to-end fashion, which allows the classification and regression branches to benefit from each other. After accomplishing the RCCN learning stage, a refinement and a fusion networks are posted to refine the discrete depth map into a higher spatial resolution. Our network achieves competitive or state-of-the-art performance on NYU Depth V2 \cite{Silberman:ECCV12}, KITTI \cite{Geiger2013IJRR}, and Make3D \cite{saxena2006learning, saxena2009make3d} benchmarks, which are three challenging datasets commonly used for~\ProblemAbbr. 

\begin{figure*}[ht!]

\begin{center}
\begin{subfigure}{1.0\textwidth}
  \begin{center}
  \includegraphics[scale=0.8]{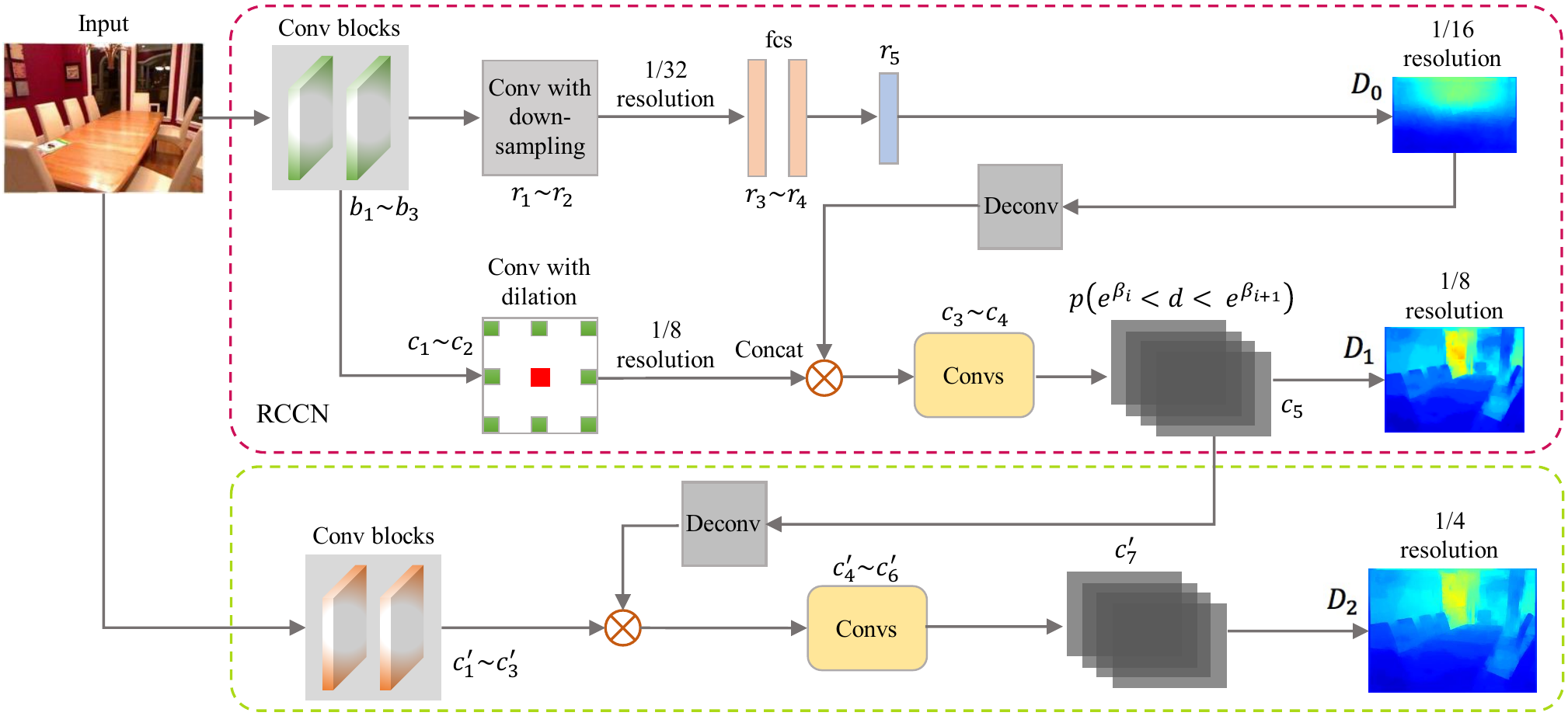}
  \end{center}
\end{subfigure}%
\caption{\small{\textbf{Network Architecture} Top: RCCN models the low spatial resolution continuous depth map and the high spatial resolution discrete depth map within a unified network to exploit the compromise between spatial and depth resolution. The regression stage achieves a full-image-view understanding and predicts a low-spatial-resolution continuous depth map. The classification stage takes the low spatial resolution continuous depth map and the convolutional features as inputs, and classifies the pixel at each spatial position to one of the pre-defined discrete depths in a high spatial resolution. Down: The refinement network takes the RGB image and those features from RCCN as inputs, and outputs a discrete depth map in a higher spatial resolution. }}
\label{fig:rcn}
\end{center}
\end{figure*}

\section{Related Work}
\label{sec:relatedwork}

Depth estimation is an important part of understanding the 3D structure of scenes from 2D images. Most prior works focused on estimating depth from stereo images by developing geometry-based algorithms \cite{scharstein2002taxonomy, forsyth2002computer} that rely on point correspondences between images and triangulation to estimate the depth. Given accurate point correspondences, depth can be estimated deterministically from stereo images. Thus, stereo depth estimation has particularly benefitted largely from the advances in local feature matching and dense optical flow estimation techniques. 

However, the geometry-based depth estimation algorithms for stereo images ignore the monocular cues in 2D images, which can also be used for depth estimation. In a seminal work \cite{saxena2006learning}, Saxena \etal learned the depth from monocular cues in 2D images via supervised learning. Since then, a variety of approaches have been proposed to exploit the monocular cues using handcrafted representations \cite{saxena2009make3d,hoiem2007recovering,ladicky2014pulling}. 
Since handcrafted features alone can only capture local information, probabilistic graphic models such as Markov Random Fields (MRFs) are often built on these features to incorporate long-range and global cues \cite{saxena2009make3d,zhuo2015indoor,liu2014discrete}. 
Another successful way to make use of global cues is the DepthTransfer method \cite{karsch2014depth} which uses GIST global scene features \cite{oliva2001modeling} to search for candidate images that are ``similar" to the input image from a database containing RGBD images. A warping procedure based on SIFT Flow \cite{liu2011sift} was then applied to the candidate image and the corresponding depth map to align them to the input image.

Given the success of DCNNs in image understanding \cite{dai2015instance, szegedy2016inception, sun2013deep, qi2016hierarchically}, some DCNNs based depth prediction frameworks have recently been proposed in recent years \cite{garg2016unsupervised, wang2015towards, godard2016unsupervised, kuznietsov2017semi}. Xie \etal \cite{xie2016deep3d} predicted the disparity map by adopting multi-level convolutional features for recovering a right view from a left-view. Garg \etal \cite{garg2016unsupervised} proposed an unsupervised framework to learn a deep depth-estimation neural network. Liu \etal \cite{liu2015deep} jointly explored the capacity of DCNNs and continuous CRF in a unified deep structured network. Moreover, Wang \etal \cite{wang2015towards} captured depth values and semantic information in a scene with DCNNs, and integrated them in a two-layer hierarchical CRF to jointly predict depth values and semantic labels. To improve efficiency, Roy and Todorovic \cite{roymonocular}  proposed the Neural Regression Forest method which allowed for parallelisable training of ``shallow" CNNs.
To further incorporate global information in DCNNs, Wang \etal \cite{wang2015designing} proposed a method for surface normal prediction, where two independent networks were learned to exploit global and local information respectively.  Eigen \etal \cite{eigen2014depth,eigen2015predicting} proposed a multi-scale network that firstly learned to predict depth at a coarse scale and then refined it using another network to produce fine-scale depth maps.
Further, Iro \etal \cite{laina2016deeper} suggested to adopt deeper network to learn better image representations towards depth estimation. 

\hh{Most recently, considering limited labeled samples and expensive human resources, some impressive unsupervised and semi-supervised \cite{garg2016unsupervised, xie2016deep3d, godard2016unsupervised, kuznietsov2017semi, zoran2015learning, chen2016single} methods were developed by posing monocular depth estimation as an image reconstruction problem. For example, Xie \etal \cite{xie2016deep3d}, Garg \etal \cite{garg2016unsupervised} and Godard \etal \cite{godard2016unsupervised} address the problem of novel view synthesis, and design reconstruction losses to estimate the disparity map by recovering a right view from a left view. In further, Kuznietsov \etal \cite{kuznietsov2017semi} incorporated extra supervision via ground truth depth in aforementioned unsupervised frameworks to improve the training. Also, Zoran \etal \cite{zoran2015learning} and Chen \etal \cite{chen2016single}} optimized depth estimation networks from selected patch pairs from images via pair-wise ranking losses, where the ordinal annotation a pair $(A, B)$ is only $d_{A}$ closer to $d_{B}$, $d_{A}$ further to $d_{B}$ and $d_{A}$ equal to $d_{B}$. 

Our RCCN method also explores global and local representations in the network for depth prediction in a supervised manner. However, our network architecture is motivated by exploiting the compromise between spatial and depth resolutions. Thus, instead of designing a stage-wise refinement procedure as done in \cite{eigen2014depth}, we introduce a cascaded structure to learn low spatial resolution but high depth resolution continuous depth via regression and high spatial resolution but low depth resolution discrete depth via classification in an end-to-end fashion. In addition, to exploit large receptive field and to alleviate information loss caused by downsampling operation, we introduce dilated convolution \cite{CP2016Deeplab} to the discrete depth estimation branch. Finally, we employ the deconvolution technique \cite{zeiler2010deconvolutional} as the bridge between the different branches to balance the feature channels from each branch.

\section{\MethodName}
\label{sec:approach}

Given an input (indoor or outdoor) image $\bf{I}$ of size $H \times W$ ($H$: height and $W$: width), we aim to predict its depth map $\mathbf{D} \in \mathbb {R}^{H \times W}$ by exploiting the compromise between spatial and depth resolution.
The key component of our approach is the proposed \emph{regression-classification cascaded network (RCCN)}. RCCN explicitly models a high spatial resolution discrete depth map $\mathbf{D}_1$ by discretizing the possible depth interval into a set of discrete values and formulates the estimation of $\mathbf{D}_1$ as a multi-class classification problem together with the regression of a low spatial resolution continuous depth map $\mathbf{D}_0$ within the deep architecture.
The obtained discrete depth map $\mathbf{D}_1$ is further refined to obtain a discrete depth map $\mathbf{D}_2$ in a higher spatial resolution via a refinement network.
We describe the detailed architecture of these two networks in Fig.~\ref{fig:rcn} to clearly show their structures and the connections.
Last but not least, the depth maps of all three scales (\ie, $\mathbf{D}_0$, $\mathbf{D}_1$, and $\mathbf{D}_2$) can be jointly considered within a fusion network, so as to achieve a continuous depth map $\tilde{\mathbf{D}}_2$\footnote{In practice, if the resolution of the depth map output by the deep model is lower than ${H \times W}$, a classic interpolation method (\eg, linear interpolation) can be used to obtain the depth map at full resolution.}.
We present the three networks in detail below.\footnote{We introduce our approach based on the network setting for our experiment on NYU Depth V2 dataset and Tab.~\ref{tab:param} provides the associated parameters, so as to facilitate the understanding of the whole approach. The network setting and parameters can be replaced by other appropriate ones.}.

\setlength\tabcolsep{2.5pt}
\begin{table}[h]
\small
\centering
\begin{tabular}{| c || c | c  c  c | c  c | c  c  c  c | c |} 
\hline
\multirow{12}{*}{\rotatebox{90}{RCCN}} & layer & $b_{1}$ & $b_{2}$ & $b_{3}$ & $r_{1}$ & $r_{2}$ & $r_{3}$ & $r_{4}$ & $r_{5}$ & \multicolumn{1}{ | c | }{De} &  \\ \cline{2-12}
 & convs & 2 & 2 & 3 & 3 & 3 & - & - & - & \multicolumn{1}{ | c | }{1} &  \\
 & chans & 64 & 128 & 256 & 512 & 512 & 2048 & 2048 & 1 & \multicolumn{1}{ | c | }{512} &  \\
 & kernel & 3 & 3 & 3 & 3 & 3 & - & - & - & \multicolumn{1}{ | c | }{4} &  \\
 & dilat & - & - & - & - & - & - & - & - & \multicolumn{1}{ | c | }{-} &  \\
 & ratio & /2 & /4 & /8 & 16/ & 32/ & - & - & /16 & \multicolumn{1}{ | c | }{/8} &  \\ \cline{2-12}
 & layer & $b_{1}$ & $b_{2}$ & $b_{3}$ & $c_{1}$ & $c_{2}$ & \multicolumn{1}{ | c | }{Co} & $c_{3}$ & $c_{4}$ & $c_{5}$ & De \\ \cline{2-12}
 & convs & 2 & 2 & 3 & 3 & 3 & \multicolumn{1}{ | c | }{-} & 1 & 1 & 1 & 1 \\
 & chans & 64 & 128 & 256 & 512 & 512 & \multicolumn{1}{ | c | }{-} & 2048 & 2048 & M & 256 \\
 & kernel & 3 & 3 & 3 & 3 & 3 & \multicolumn{1}{ | c | }{-} & 3 & 1 & 1 & 4 \\
 & dilat & - & - & - & - & 2 & \multicolumn{1}{ | c | }{-} & - & - & - & - \\
 & ratio & /2 & /4 & /8 & /8 & /8 & \multicolumn{1}{ | c | }{/8} & /8 & /8 & /8 & /4 \\
 \hline
 \multirow{5}{*}{\rotatebox{90}{Refinement}} 
 & layer & $c'_{1}$ & $c'_{2}$ & $c'_{3}$  &  &  & \multicolumn{1}{ | c | }{Co} & $c'_{4}$ & $c'_{5}$ & $c'_{6}$ & $c'_{7}$ \\ \cline{2-12}
  & convs & 2 & 2 & 3 &  &  & \multicolumn{1}{ | c | }{-} & 1 & 1 & 1 & 1 \\
 & chans & 64 & 128 & 256 &  &  & \multicolumn{1}{ | c | }{-} & 1024 & 1024 & 1024 & M \\
 & kernel & 3 & 3 & 3 &  &  & \multicolumn{1}{ | c | }{-} & 3 & 3 & 1 & 1 \\
  & ratio & /4 & /4 & /4 & /4 & /4 & \multicolumn{1}{ | c | }{/4} & /4 & /4 & /4 & /4 \\
 \hline

\end{tabular}
\caption{\small{\textbf{Network Parameters.} Parameters and neurons of the proposed network for NYU Depth V2 dataset based on VGG. $b_{i}$: layer shared by the two branches in RCCN. $r_{i}$/$c_{i}$: layer of the continuous/discrete depth estimation branch. $c'_{i}$: layer of the refinement network. Co: concatenation layer.  De: deconvolutional layer. M: the number of pre-define sub-intervals.}} 
\label{tab:param}
\end{table}

\subsection{Regression-Classification Cascaded Network}

The Regression-Classification Cascaded Network (RCCN) is a joint regressor-classifier. It serves as a two-tier estimator that simultaneously predicts the initial continuous depth map $\mathbf{D}_0$ and the discrete depth map $\mathbf{D}_1$. We choose to adopt a two-stage cascaded network for modeling and implementing the regression-classification network,
aiming to exploit the compromise between low spatial resolution continuous depth and high spatial resolution discrete depth. We also incorporate global scene information from the entire image and structural and contextual information in a large receptive field.
\newline

\noindent \textbf{Regressing Continuous Depth:} In this stage, the network aims to predict the low spatial resolution continuous depth map  $\mathbf{D}_0$ from a global understanding of the entire image, by abstracting a representation feature vector from the whole image field of view. From such a representation, we learn specific non-linear functions for all the pixels located at a pre-defined resolution ($h \times w$).

To this end, on top of the shared convolutional layers ($b_{1}$ to $b_{3}$), additional convolutional layers ($r_{1}$ and $r_{2}$) and max-pooling layers with downsampling are used to obtain deeper convolutional features at a coarse resolution. Then, after the pass of two fully-connected ($fc$) layers ($r_{3}$ and $r_{4}$), the feature vector will contain high-level information of the whole input image. Followed by a third $fc$ layer ($r_{5}$) with $h \times w$ outputs, each output represents the depth value of a spatial location within the pre-defined resolution, and connects to all the vector elements from the last layer, implying a global understanding of the entire image.


This stage is supervised by manually labeled continuous depth values over the whole input image in stride 8 via root mean squared error in $\log$ space (RMSE$_{log}$). More specifically, we reshape the $hw$-dimensional output vector to a $h \times w$ map to obtain the predicted depth map $\tilde{D}^{\prime}$ at this stage to compare it with the target depth map $D^{\prime}$. 

It should be noted that, the RMSE$_{log}$ loss uses a $log$ function to down-weight the losses in regions with large depth values, which is commonly used as an evaluation metric. From a statistical view, let us consider the generation of data following $y = exp(log(x) + N(0,1))$ with $x$ uniformed distributed. It is easy to observe that the noise variance of $y$ is larger when $x$ is larger, implying that the observed depth value has a larger noise variance when its ground truth is larger. Hence, without $log$, large depth values would induce an over-strengthened influence on the training process, which is not expected. It also motivates the SID method (instead of uniform quantization) for classification as below, which quantizes depth values with increasing intervals and whose advantage is quantitatively evaluated.
\newline

\noindent \textbf{Categorizing Approximate Depth:} This stage categorizes each pixel to one of the pre-defined discrete depths in a higher spatial resolution, by taking the shared convolutional features and the previous-stage depth map as inputs.


\begin{figure}[ht!]

\begin{center}
\begin{subfigure}{0.48\textwidth}
  \begin{center}
  \includegraphics[scale=0.45]{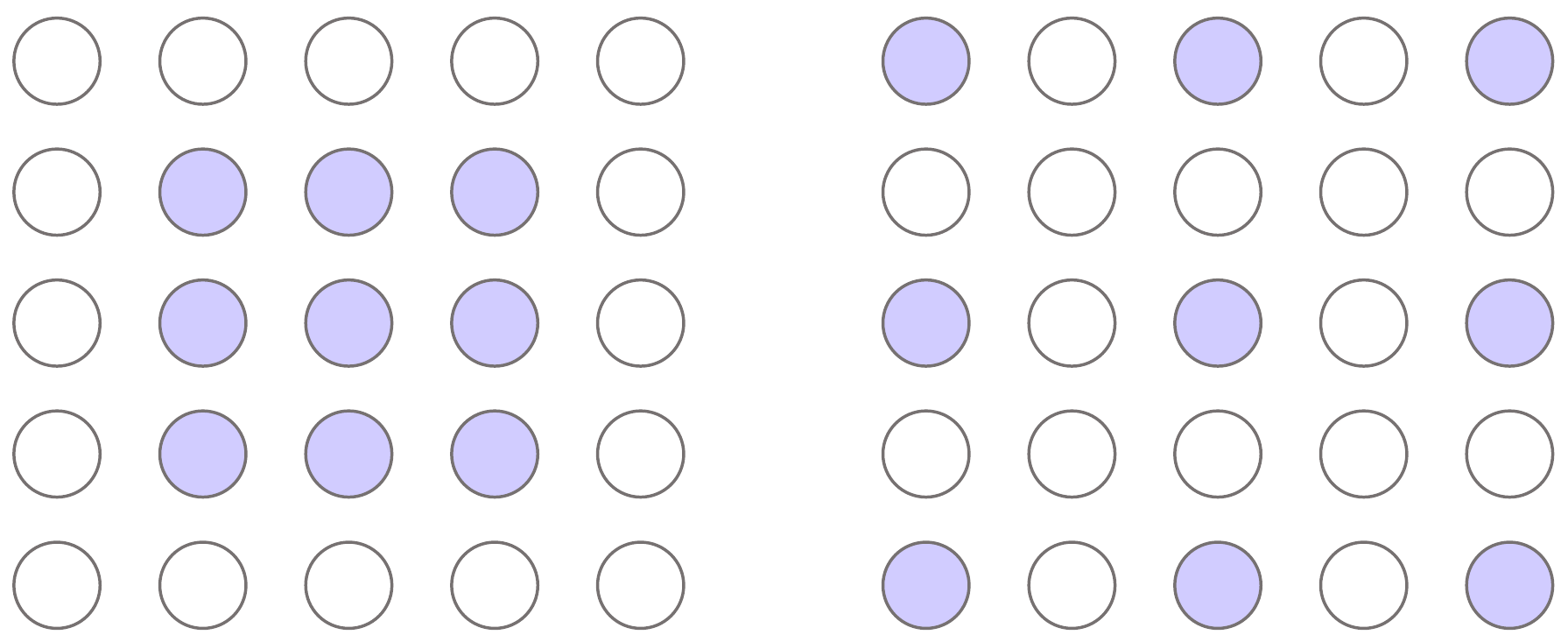}
  \end{center}
\end{subfigure}%
\caption{\small{\textbf{Dilated Convolution.} Left: regular convolutional layer ($c_{1}$). Right: dilated convolutional layer ($c_{2}$). }}
\label{fig:dilat}
\end{center}
\end{figure}

Specifically, in order to better exploit the compromise between spatial and depth resolution as well as the geometric contexts and physical properties of the image, we adopt a cascaded structure in which the previous-stage depth map is fed into the classification network. The previous-stage depth map is spatially coarse (low spatial resolution) but provides a global-field-of-view understanding of the input image. 
The shared convolutional features contain finer spatial information. 
Therefore, on top of the shared convolutional layers, additional convolutional layers ($c_{1}$ and $c_{2}$) are used to obtain local structural and contextual information. In contrast to the regression branch, we skip the subsampling operation in the max-pooling layers and employ the dilated convolution technique ($c_{2}$) \cite{CP2016Deeplab}, which introduces zeros to increase the convolution field to exploit large receptive field information in the fine resolution feature map, as shown in Fig.~\ref{fig:dilat}. The predicted continuous depth map is simultaneously deconvoluted to multi-channel feature maps with the same spatial resolution as $c_{2}$. The deconvolution to multi-channel features is an important component to balance the features from different branches. Followed by a concatenation layer, three extra convolutional layers ($c_{3}$ to $c_{5}$) are applied to learn a richer representation and model the probabilities of the depth sub-intervals that each pixel belongs to.


This stage is supervised by the pre-defined discrete depths $D$ like semantic labels in segmentation tasks. We minimize the multinomial logistic loss to learn the network parameters.

As for discretization strategies, uniform discretization (UD) is a common way to obtain a set of representative values from a depth interval $[a, b]$. However, considering the facts that the importance of a fixed interval (\eg, $1m$) decreases when the depth ranges from small to large, we propose to use the following spacing-increasing discretization (SID) strategy so that the learned model pays more attention to estimating relatively small depths:
\begin{equation}
\begin{split}
 \text{UD: \quad}    l_{j}  & = a + (b - a)*j/K
 \\
 \text{SID: \quad}    l_{j}  & = e^{\log(a) + \frac{\log(b/a)*j}{K}}
 \end{split}
 \label{eq:dis}
\end{equation}
where $K$ is a pre-define sub-interval number.

 \setlength\tabcolsep{9pt}
\begin{table*}[h!]
\normalsize
\centering
\begin{tabular}{ c || c | c | c || c | c | c | c }
\hline
\multirow{ 2 }{*}{Method}  & \multicolumn{3}{   |c || }{ higher is better }	 &  \multicolumn{4}{   c  }{ lower is better } \\ \cline{2-8}
 & $\delta < 1.25$ & $\delta < 1.25^{2}$ & $\delta < 1.25^{3}$ & Abs Rel & Squa Rel & $\text{RMSE}$ & $\text{RMSE}_{log}$  \\
\hline\hline
Make3D \cite{saxena2009make3d} & 0.601 & 0.820 & 0.926 & 0.280 & 3.012 & 8.734 & 0.361 \\
Eigen \etal \cite{eigen2014depth} &  0.692 & 0.899 & 0.967 & 0.190 & 1.515 & 7.156 & 0.270 \\
Liu \etal \cite{liu2016learning} &   0.647 & 0.882 & 0.961 & 0.217 & 1.841 & 6.986 & 0.289 \\
LRC (CS + K) \cite{godard2016unsupervised} &  0.861 & 0.949 & 0.976 & 0.114 & 0.898 & 4.935 & 0.206 \\
Kuznietsov \etal \cite{kuznietsov2017semi} &  0.862 & 0.960 & 0.986 & 0.113 & 0.741 & 4.621 & 0.189 \\
\hline
RCCN-VGG &  0.870 & 0.970 & 0.993 & 0.110 &  0.620 & 4.029 & 0.160 \\
RCCN-VGG$^{\dagger}$ &  0.886 & 0.975 & 0.994 & 0.105 &  0.540 & 3.903 & 0.154 \\
RCCN-ResNet$^{\dagger}$ &  0.911 & 0.979 & 0.993 & 0.084 &  0.386 & 3.072 & 0.136 \\
\hline
\end{tabular}
\caption{\small{\textbf{Performance on KITTI test set.} All the scores are evaluated on Eigen \etal \cite{eigen2014depth} test split. RCCN-VGG/ResNet: RCCN with backbones of VGG or ResNet. $\dagger$: RCCN with post refinement processes.}} 
\label{tab:kitti}
\end{table*}

\subsection{Learning}
Let $\Gamma = \zeta(\Omega, \Upsilon, I)$ denote the feature vector containing $m$ elements outputting from $r_{4}$, where $\Omega$ is shared parameters in the \emph{conv} blocks $b_{1} \sim b_{3}$, and $\Upsilon$ is the parameters in $r_{1} \sim r_{4}$. $\Lambda =  \xi(\Omega, \Upsilon, \Xi, \Theta_{r}, I)$ are the feature maps with size $W \times H \times C $ outputting from $c_{4}$, where $\Xi$ is parameters in $c_{1} \sim c_{4}$, and the deconvolutional layer between $r_{5}$ and $c_{3}$, $\Theta_{r}$ are parameters of $r_{5}$.  $\tilde{D}^{\prime} = \phi (\Gamma, \Theta)$ of size $W^{\prime} \times H^{\prime}$ denotes predicted depth map in the regression stage, and $Y = \psi (\Lambda, \Theta)$ of size $W \times H \times K$ denotes category outputs for each spatial locations, where $\Theta = ( \Theta_{r}; \Theta_{c}) = ( \theta_{0}, \theta_{1}, ..., \theta_{W^{\prime}*H^{\prime} - 1}; \theta_{W^{\prime}*H^{\prime}}, ..., \theta_{W^{\prime}*H^{\prime} + K - 1})$ are parameters of $r_{5}$ and $c_{5}$. The loss function for our RCCN will take the form as:

\begin{equation}
\begin{split}
\mathcal{L} (\Gamma, \Lambda, \Theta) &= \mathcal{L}_{r} (\Gamma, \Theta_{r}) + \mathcal{L}_{c} (\Lambda, \Theta_{c}) \\
&= \frac{1}{\mathcal{N}_r} \sum_{w=0}^{W^{\prime} - 1}\sum_{h=0}^{H^{\prime}-1}  \Phi_{r} (\Gamma, \Theta_{r}, w, h) \\
&\quad + \frac{1}{\mathcal{N}_c} \sum_{w=0}^{W - 1}\sum_{h=0}^{H - 1} \Phi_{c} (\Lambda, \Theta_{c}, w, h), 
\\
\Phi_{r} (\Gamma, \Theta_{r}, w, h) &= || \tilde{d}_{(w, h)}^{\prime} - d_{(w,h)}^{\prime} {||}^{2}, \\
\Phi_{c} (\Lambda, \Theta_{c}, w, h) &= - \sum_{k=0}^{K - 1} 1\{ d_{(w, h)} = k\} \log (\mathcal{P}_{(w, h)}^{k}), \\
\mathcal{P}_{(w, h)}^{k} &= P(\tilde{d}_{(w,h)} = k | \Lambda, \Theta_{c}), \\ 
\end{split}
\label{eq:loss}
\end{equation}
where $1\{ \cdot \}$ is a indicator function, so that $1 \{ true \} = 1$, and $1\{ false \} = 0$, $\mathcal{N}_{r} = W^{\prime} \times H^{\prime}$, $\mathcal{N}_{c} = W \times H$, $\tilde{d}_{(w,h)}^{\prime} \in \tilde{D}^{\prime}$, $d_{(w,h)}^{\prime} \in D^{\prime}$, $K$ is the number of sub-intervals, $d_{(w, h)} \in D$ is the ground-truth discrete depth value in spatial location $(w, h)$. The \emph{softmax} regression for the classification stage computes $\mathcal{P}_{(w, h)}^{k}$ as:
\begin{equation}
\begin{split}
\mathcal{P}_{(w, h)}^{k} &= \frac{e^{y_{(w,h,k)}}}{\sum_{i=0}^{K-1}e^{y_{(w,h,i)}}}, \\
\end{split}
\label{eq:prob}
\end{equation}
where $y_{(w,h,i)} = \theta_{W^{\prime}*H^{\prime}+i}^{T}\gamma_{(w, h)}$, and $\gamma_{(w, h)} \in \Gamma$.

To minimize $\mathcal{L} (\Gamma, \Lambda, \Theta)$, taking derivate with respect to $\theta_{i}$, we can obtain the gradient as: 
\begin{equation}
\begin{split}
\frac{\partial{\mathcal{L} (\Gamma, \Lambda, \Theta)}}{\partial{\theta_{i}}} &= \frac{\partial{ \mathcal{L}_{r} (\Gamma, \Theta_{r})}}{\partial{\theta_{i}}} + \frac{\partial{\mathcal{L}_{c} (\Lambda, \Theta_{c})}}{\partial{\theta_{i}}} \\
&=  \frac{1}{\mathcal{N}_r} \sum_{w=0}^{W^{\prime} - 1}\sum_{h=0}^{H^{\prime}-1} \frac{\partial{ \Phi_{r} (\Gamma, \Theta_{r}, w, h)}}{\partial{\theta_{i}}}
\\
&\quad + \frac{1}{\mathcal{N}_c} \sum_{w=0}^{W - 1}\sum_{h=0}^{H - 1} \frac{\partial{\Phi_{c} (\Lambda, \Theta_{c}, w, h)}}{\partial{\theta_{i}}}.
\\
\end{split}
\label{eq:gradient}
\end{equation}
We compute $\frac{\partial{ \Phi_{r} (\Gamma, \Theta_{r}, w, h)}}{\partial{\theta_{i}}}$ and $\frac{\partial{\Phi_{c} (\Lambda, \Theta_{c}, w, h)}}{\partial{\theta_{i}}}$ respectively as follow: 
\begin{equation}
\begin{split}
\frac{\partial{ \Phi_{r} (\Gamma, \Theta_{r}, w, h)}}{\partial{\theta_{i}}} &= 1\{i < W^{\prime}*H^{\prime}\}1\{ i = w*H^{\prime}+h\} \\
& \quad \times 2(\tilde{d}_{(w, h)}^{\prime} - d_{(w,h)}^{\prime})\Gamma,
\\
\frac{\partial{ \Phi_{c} (\Lambda, \Theta_{c}, w, h)}}{\partial{\theta_{i}}} &= 1\{i < W^{\prime}*H^{\prime}\}\frac{\partial{ \Phi_{c} (\Lambda, \Theta_{c}, w, h)}}{\partial{\theta_{i}}} \\
& \quad + 1\{i \ge W^{\prime}*H^{\prime}\} \\
& \quad \times \lambda_{(w,h)}(\mathcal{P}_{(w,h)}^{i} - 1\{\tilde{d}_{(w,h)} = i\}),
\\
\end{split}
\label{eq:gradient1}
\end{equation}
where $i \in \{0, 1, ..., W^{\prime}*H^{\prime} + K - 1\}$ and $\frac{\partial{ \Phi_{c} (\Lambda, \Theta_{c}, w, h)}}{\partial{\theta_{i}}}$ can be computed via chain rules and backpropagation when $i < W^{\prime}*H^{\prime}$ ($\Lambda$ also parameterized by $\Theta_{r}$).


%
%
%
\subsection{Post Refinement}
\subsubsection{Refinement network}
By taking the input RGB image and the features of the last classification branches of RCCN as inputs, the refinement network incorporate multi-scale features (implying different receptive field sizes) and refines the discrete depth map into a higher spatial resolution.

Via those convolution blocks ($c'_{1}$ to $c'_{3}$), we obtain features at a $1/4$ resolution of the original image (a relative small receptive field). Similar to the second stage in RCCN, the features ($c_{5}$) are deconvoluted into the same resolution of $c'_{3}$ with multi-channels outputs. Then, convolutional layers ($c'_{4}$ to $c'_{8}$) are applied on these two scales of features to obtain the refined discrete depth map. The supervised information in this refinement network is the same as the one in the second stage of RCCN, except for the higher spatial resolution. We initialize the trained parameters of $c'_{1} \sim c'_{3}$ using those of $b_{1}  \sim  b_{3}$ in our experiments\footnote{Note that $c'_{1} \sim c'_{3}$ are independent from  $b_{1}  \sim  b_{3}$.}.
\newline

\subsubsection{Fusion Network}
\label{subsec:comb}
The depth map from the refinement network already incorporates multi-scale features in multi-scale receptive fields, and on average achieves a more accurate depth map estimation than those two predicted maps of RCCN, but not for all individual pixels. There are are two possible reasons for such a phenomenon: (i) the refined depth map is still in discrete space; or (ii) depth estimation for some objects (especially for large objects) depends more on information from large receptive fields, and features from small receptive fields may introduce some noises. To address this, we integrate the depth maps from all three scales within the fusion network, which just consists of a few convolutional layers in our experiments. Also, the supervised information in the fusion network is the same as that in the first stage of RCCN except for the higher spatial resolution.

\setlength\tabcolsep{9pt}
\begin{table*}[h]
\normalsize
\centering
\begin{tabular}{ l || c | c || c | c | c || c | c } 
\hline
 & {R} & {C} & {RRCN}  & {RCCN}  & {CCCN} & {$D_{0}$} & {RCCN-$D_{0}$}\\ 
 \hline\hline
$\delta < 1.25$ & 0.732 & 0.806 & 0.762 & 0.852  & 0.835 & 0.741 & 0.750 \\
$\delta < 1.25^{2}$ & 0.905 & 0.947 & 0.928 & 0.963  & 0.950 & 0.913 & 0.919 \\
$\delta < 1.25^{3}$ & 0.951 & 0.985 & 0.968 & 0.990  & 0.982 & 0.956 & 0.960 \\
\hline
Abs Rel & 0.172 & 0.142 & 0.162 & 0.123 & 0.132 & 0.168 & 0.162 \\
Squa Rel & 1.105 & 0.892 & 1.060 & 0.763  & 0.893 & 1.092 & 1.071 \\
$\text{RMSE}$ & 5.829 & 4.711 & 5.105 & 4.235  & 5.102 & 5.476 & 5.265 \\
$\text{RMSE}_{log}$ & 0.282 & 0.198 & 0.235 & 0.174  & 0.208 & 0.270 & 0.247 \\
 \hline

\end{tabular}
\caption{\small{\textbf{Variants of RCCN on KITTI dataset.}  R: Directly learning continuous depth in a fine scale via regression. C: Directly learning discrete depth in a fine scale via classification with proposed SID strategy. RRCN: Learning continuous depth via a cascaded regression-regression structure. RCCN: The proposed method. CCCN: A classification-classification cascade network. Note that, higher is better in top table, while lower is better in bottom table.}} 
\label{tab:varient}
\end{table*}

\begin{figure*}[ht!]

\begin{center}
\begin{subfigure}{0.48\textwidth}
  \begin{center}
  \includegraphics[scale=0.5]{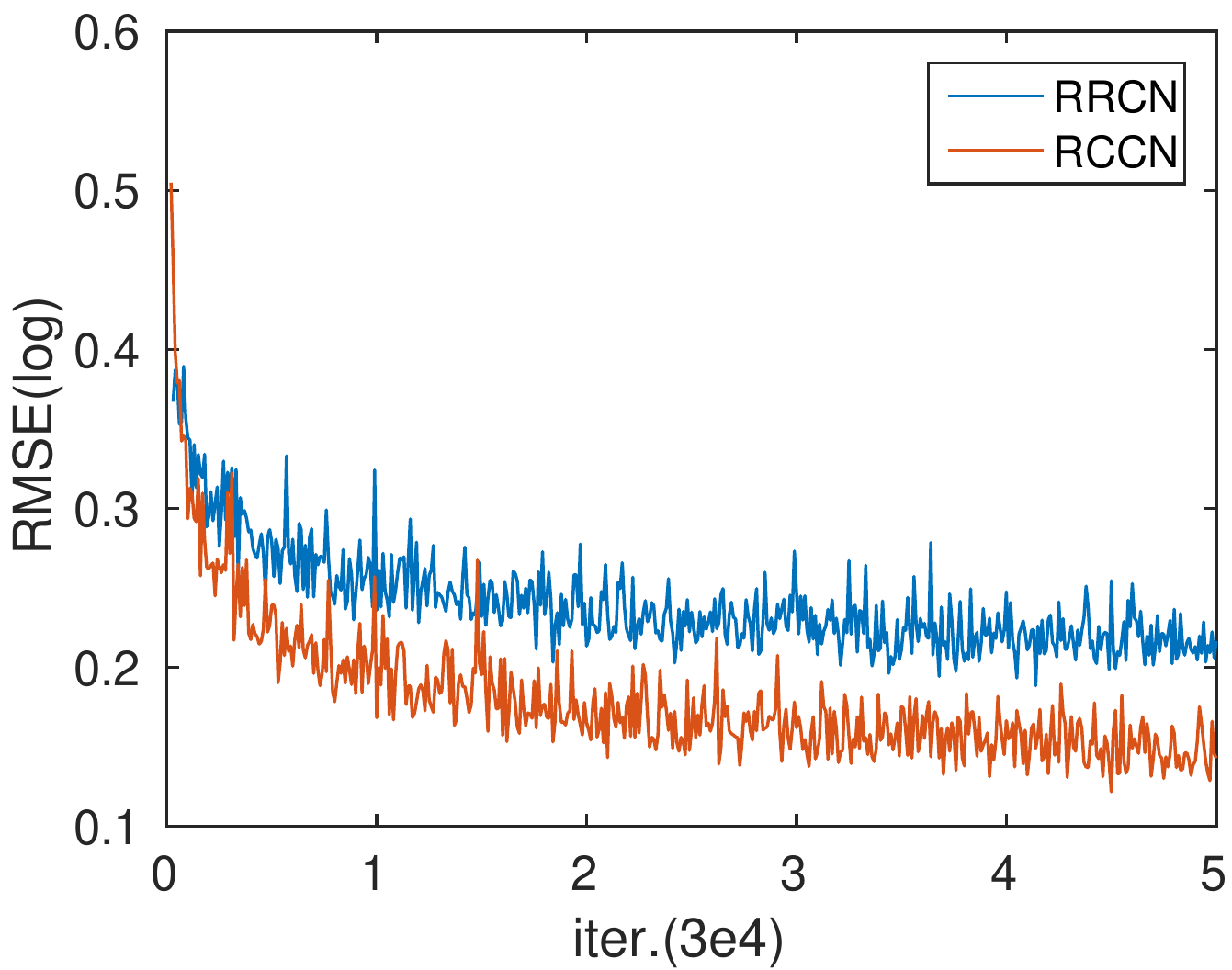}
  \end{center}
\end{subfigure}%
\begin{subfigure}{0.48\textwidth}
  \begin{center}
  \includegraphics[scale=0.5]{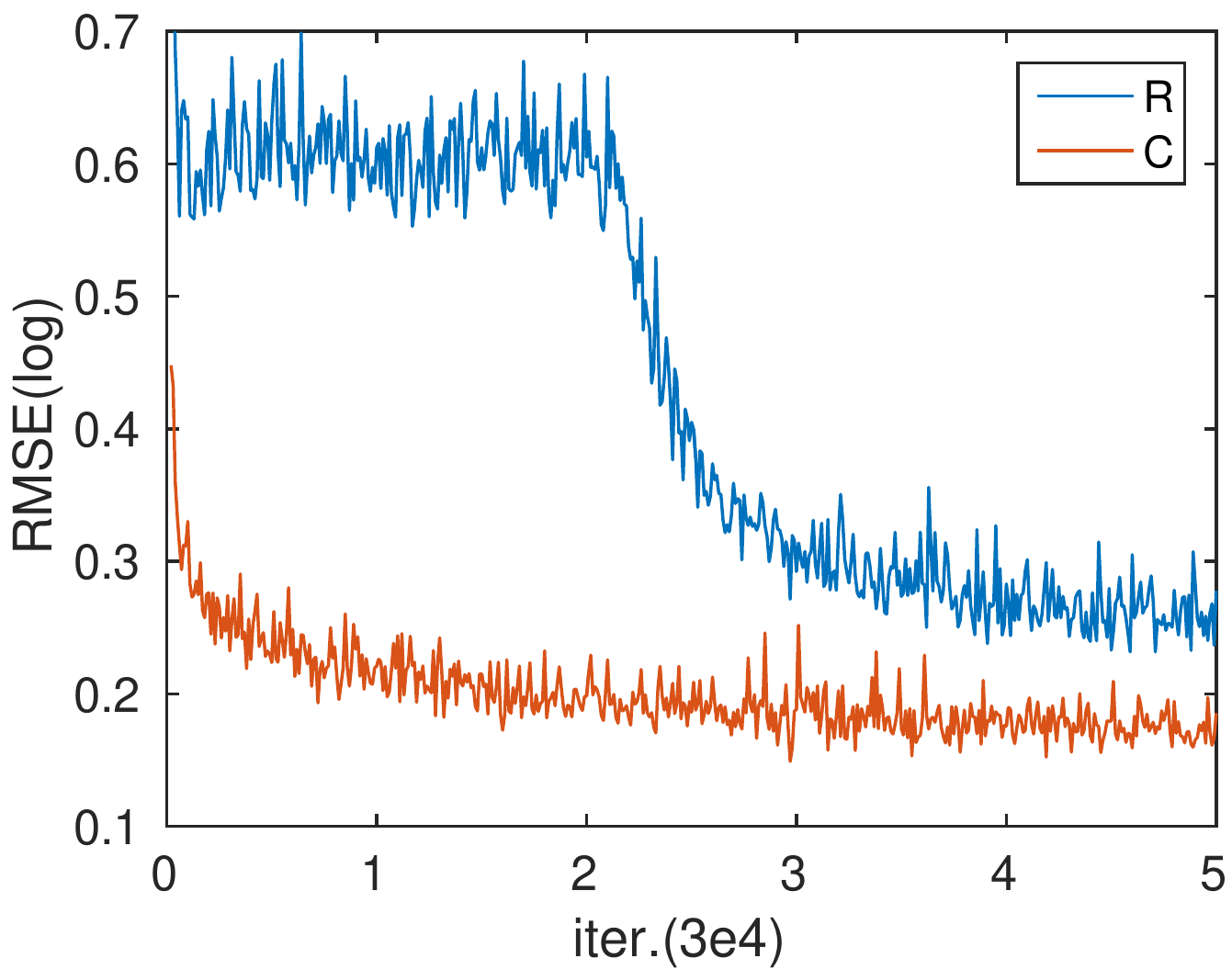}
  \end{center}
\end{subfigure}%
\caption{\small{\textbf{Training Error on KITTI dataset.}  Left:  Training errors of RCCN and RRCN.  Right: Training error of R and C. From the two illustrations, our classification strategy (C) for depth estimation can make the neural network converge to a better solution rather than regression strategy (R).}}
\label{fig:loss}
\end{center}
\end{figure*}


\section{Experiments}
\label{sec:experiments}
To validate the compromise principle and demonstrate the effectiveness of RCCN, here we present a number of experiments examining different aspects of the approach. After introducing the common experimental settings, we evaluate our methods on three challenging datasets, \ie \emph{NYU Depth V2} \cite{Silberman:ECCV12}, \emph{KITTI} \cite{Geiger2013IJRR}, and \emph{Make3D} \cite{saxena2006learning, saxena2009make3d}, via the error metrics using in previous works.

 \setlength\tabcolsep{9pt}
\begin{table*}[h]
\normalsize
\centering
\begin{tabular}{ c || c | c | c || c | c | c | c  }
\hline
\multirow{ 2 }{*}{Method} & \multicolumn{3}{   |c || }{ higher is better }	 &  \multicolumn{4}{   c  }{ lower is better } \\ \cline{2-8}
 & $\delta < 1.25$ & $\delta < 1.25^{2}$ & $\delta < 1.25^{3}$ & Abs Rel & Squa Rel & $\text{RMSE}$ & $\text{RMSE}_{log}$ \\
\hline\hline
Make3D \cite{saxena2009make3d} & 0.447 & 0.745 & 0.897 & 0.349 & 0.492 & 1.214  & 0.409  \\
DepthTransfer \cite{karsch2014depth} & 0.460 & 0.742 & 0.893 & 0.350 & 0.539 & 1.1 & 0.378  \\
Liu \etal \cite{liu2014discrete} & 0.475 & 0.770 & 0.911 & 0.335 & 0.442 & 1.06 & 0.362  \\
Ladicky \etal \cite{ladicky2014pulling} & 0.542 & 0.829 & 0.941 & - & - & - & -  \\
Li \etal \cite{li2015depth} & 0.621 & 0.886 & 0.968 & 0.232 & - & 0.821 & -  \\
Wang \etal \cite{wang2015towards} & 0.605 & 0.890 & 0.970 & 0.220 & 0.210 & 0.745 & 0.262  \\
Roy \etal \cite{roymonocular} & - & - & - & 0.187 & - & 0.744 & -  \\
Liu \etal \cite{liu2016learning} & 0.650 & 0.906 & 0.976 & 0.213 & - & 0.759 & -  \\
Eigen \etal \cite{eigen2015predicting} & 0.769 & 0.950 & 0.988 & 0.158 & 0.121 & 0.641 & 0.214 \\
Chakrabarti \etal \cite{chakrabarti2016depth} & 0.806 & 0.958 & 0.987 & 0.149 & 0.118 & 0.620 & 0.205  \\
Laina \etal \cite{laina2016deeper} & 0.629 & 0.889 & 0.971 & 0.194 & - & 0.790 & -  \\
Xu \etal \cite{xu2017multi} & 0.636 & 0.896 & 0.972 & 0.193 & - & 0.792 & - \\
Li \etal \cite{Li_2017_ICCV} & 0.789 & 0.955 & 0.988 & 0.152 & - & 0.611 & -  \\
Laina \etal \cite{laina2016deeper}$^\dag$ & 0.811 & 0.953 & 0.988 & 0.127 & - & 0.573 & 0.195  \\
Li \etal \cite{Li_2017_ICCV}$^\dag$ & 0.788 & 0.958 & 0.991 & 0.143 & - & 0.635 & -  \\
Xu \etal \cite{xu2017multi}$^\dag$ & 0.811 & 0.954 & 0.987 & 0.121 & - & 0.586 & - \\
\hline
RCCN-VGG & 0.753 & 0.937 & 0.983 & 0.165 & 0.138 &  0.607 & 0.213  \\
RCCN-VGG$^{\dagger}$ & 0.765 & 0.950 & 0.991 & 0.160 & 0.131 &  0.586 & 0.204  \\
RCCN-ResNet$^{\dagger}$ & 0.807 & 0.957 & 0.992 & 0.136 & 0.116 &  0.564 & 0.199 \\
\hline
\end{tabular}
\caption{\small{\textbf{Performance on NYU Depth v2 test set.} All the scores are evaluated on offical test split. RCCN-VGG/ResNet: RCCN with backbones of VGG or ResNet. $\dagger$: RCCN with post refinement processes.}} 
\label{tab:nyuv2}
\end{table*}

\begin{figure*}[ht!]

\begin{center}

\begin{subfigure}{0.96\textwidth}
  \begin{center}
  \includegraphics[scale=0.96]{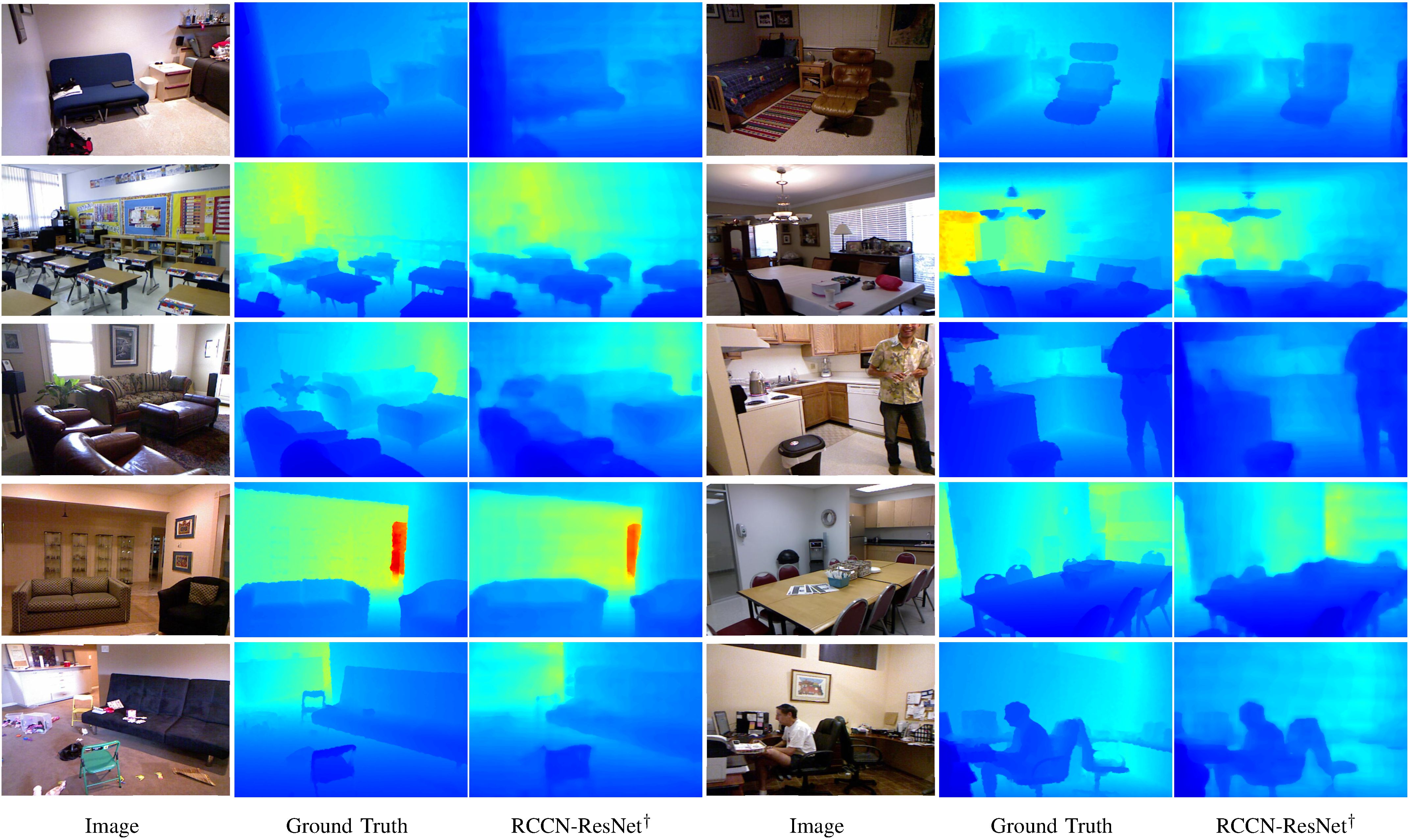}
  \end{center}
\end{subfigure}%

\caption{\small{\textbf{Depth Prediction on NYU Depth v2.}}}
\label{fig:nyu}
\end{center}
\end{figure*}

\subsection{Experimental Setting}


\subsubsection{Implementation Details}
In order to fairly compare the proposed method with the current state-of-the-art methods,
our method adopts both VGG-16 \cite{simonyan2014very} and ResNet-101 \cite{he2016deep} as our backbone. We initialise the parameters of RCCN in base convolutional layers via the pre-trained classification model on ILSVRC \cite{ILSVRC15}. The training procedures for RCCN-VGG and RCCN-ResNet are a little different. For RCCN-VGG, we directly train our network follows a polynomial decay with a base learning rate of $0.0001$, the power of $0.9$, the momentum of $0.9$, and the weight decay of $0.0005$. However, for RCCN-ResNet, we find that directly optimize our network with a large base learning rate resulting an unexpected divergence after some iterations, and a small base learning rate resulting slow convergence rate. To speed up training, we fixed all the parameters in the base convolutional layers, and first train the regression stage for a few iterations, and then train the two stage for some other iterations with a base learning rate of $0.001$. And finally, we optimize all the parameters together in RCCN-ResNet with a base learning rate of $0.0001$. Further, the networks in post refinement stages are independently trained with fixed parameters of RCCN. The proposed method is implemented via a public deep learning paltform \emph{Caffe} \cite{jia2014caffe}, and trained on 4 TITAN X GPUs with 12GB of memory per GPU with batch size of 4.
\newline


\subsubsection{Data Augmentation}
Following previous works \cite{eigen2014depth, laina2016deeper}, we employ some  data augmentation techniques to prevent overfitting and to learn a better model in the training process, including: (i) \emph{Random Cropping:} we randomly crop rectangles with predefined sizes from the original image. (ii) \emph{Flipping:} we randomly flip the original image horizontally. (iii) \emph{Scaling:} we randomly resize the original image by a scale factor belongs to the interval of $[0.75, 1.25]$, and normalize the associated depth map with the corresponding scales. (iv) \emph{Rotation:} we randomly rotate the input image with the degree of $[-10^{\circ}, 10^{\circ}]$.
\newline

\subsubsection{Evaluation Metrics}
Below are the list of depth error metrics based on which the quantitative evaluation is performed: 

\begin{equation}
\begin{split}
& \text{Accuracy}:  \frac{1}{T} \sum_{i \in T} 1\{\delta = max(\frac{d_{i}}{\hat{d}_{i}}, \frac{\hat{d}_{i}}{d_{i}}) < thr \} \\
& \text{Abs Rel}: \frac{1}{ N }\sum_{i \in N}\frac{|d_{i} - d_{i}^{*}|}{d_{i}^{*}} \\
& \text{Squa Rel}: \frac{1}{ N }\sum_{i \in N}  \frac{\| d_{i} - d_{i}^{*} \|^2}  {d_{i}^{*}} \\
& \text{RMSE}: \sqrt{\frac{1}{N }\sum_{i \in N} \Vert d_{i} - d_{i}^{*} \Vert^{2}} \\
& \text{RMSE$_{log}$}: \sqrt{\frac{1}{N }\sum_{i \in N} \Vert \log (d_{i}) - \log(d_{i}^{*}) \Vert^{2}} \\
& \text{Ave $\log_{10}$}: \frac{1}{N }\sum_{i \in N} | \log_{10} (d_{i}) - \log_{10}(d_{i}^{*}) | \\
\end{split}
\label{eq:err}
\end{equation}

 \setlength\tabcolsep{9pt}
\begin{table*}[h]
\normalsize
\centering
\begin{tabular}{ c || c | c | c || c | c | c }
\hline
\multirow{ 2 }{*}{Algorithm} & \multicolumn{3}{  | c || }{ C1 error } & \multicolumn{3}{  | c  }{ C2 error } \\ \cline{2-7}
 & Abs Rel & Ave $\log_{10}$ & RMSE  & Abs Rel & Ave $\log_{10}$ & RMSE \\
\hline\hline
Make3D \cite{saxena2009make3d} & - & - & - & 0.370 & 0.187 & -\\
Liu \etal \cite{liu2010single} & - & - & - & 0.379 & 0.148 & -\\
DepthTransfer \cite{karsch2014depth} & 0.355 & 0.127 & 9.20 & 0.361 & 0.148 & 15.10 \\
Liu \etal \cite{liu2014discrete} & 0.335 & 0.137 & 9.49 & 0.338 & 0.134 & 12.60 \\
Li \etal \cite{li2015depth} & 0.278 & 0.092 & 7.12 & 0.279 & 0.102 & 10.27\\
Liu \etal \cite{liu2016learning} & 0.287 & 0.109 & 7.36 & 0.287 & 0.122 & 14.09\\
Roy \etal \cite{roymonocular} & - & - & - & 0.260 & 0.119 & 12.40\\
Laina \etal \cite{laina2016deeper} & 0.176 & 0.072 & 4.46 & - & - & -\\
LRC-Deep3D \cite{xie2016deep3d} & 1.000 & 2.527 & 19.11 & - & - & -\\
LRC \cite{godard2016unsupervised} & 0.443 & 0.156 & 11.513 & - & - & -\\
Kuznietsov \etal \cite{kuznietsov2017semi} & 0.421 & 0.190 & 8.24 & - & - & -\\
Xu \etal \cite{xu2017multi} & 0.184 & 0.065 & 4.38 & 0.198 & 4.53 & 8.56 \\
\hline
RCCN-VGG$^{\dagger}$ & 0.252 & 0.104 & 8.82  & 0.255 & 0.106 & 11.57 \\
RCCN-ResNet$^{\dagger}$ & 0.189 &0.082 & 5.57 & 0.192 & 0.088 & 9.34 \\
\hline
\end{tabular}
\caption{\small{\textbf{Performance on Make3D test set.}} LRC-Deep3D \cite{xie2016deep3d} is adopting LRC \cite{godard2016unsupervised} on Deep3D model \cite{xie2016deep3d}.} 
\label{tab:make3d}
\end{table*}

\begin{figure*}[ht!]

\begin{center}

\begin{subfigure}{0.96\textwidth}
  \begin{center}
  \includegraphics[scale=0.94]{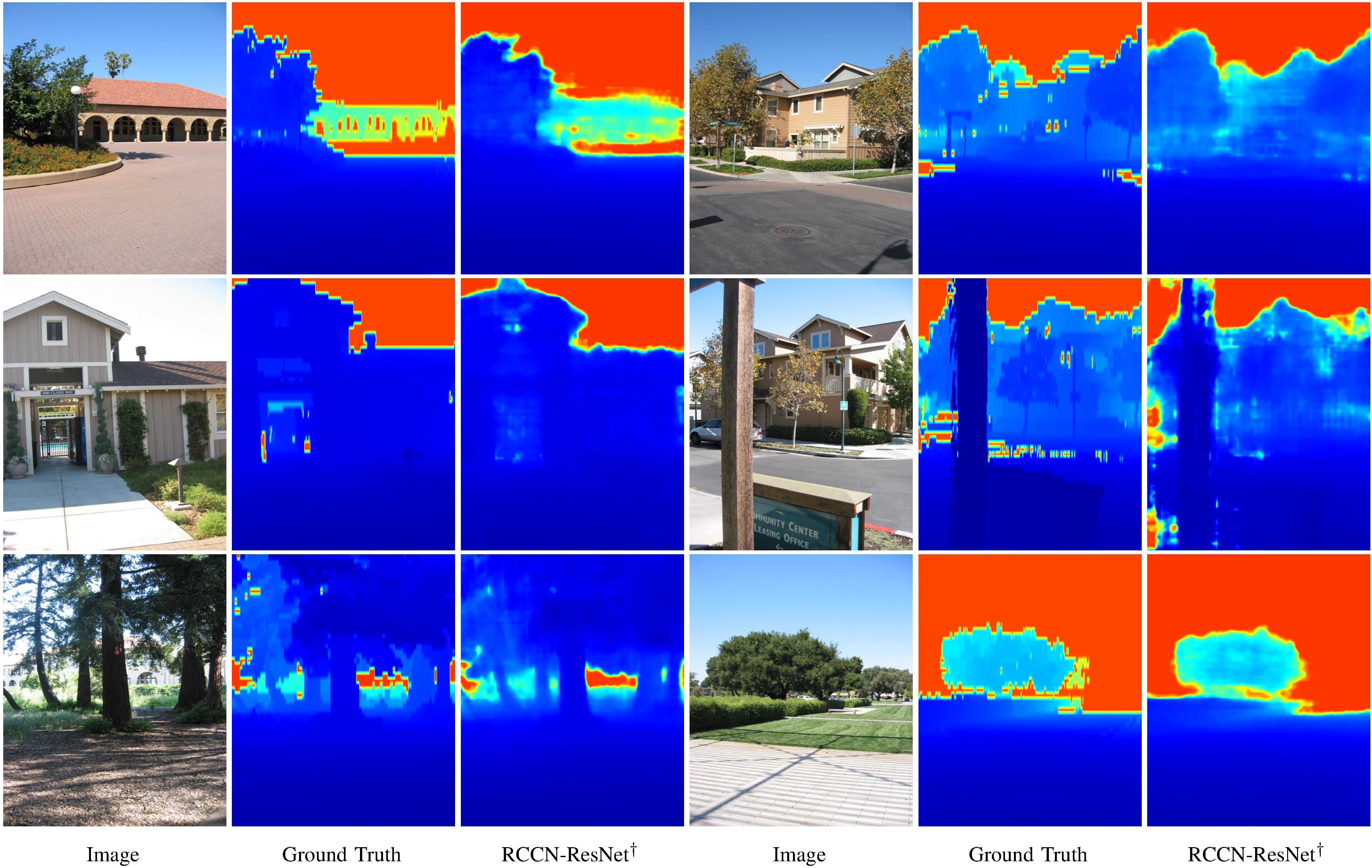}
  \end{center}
\end{subfigure}%

\caption{\small{\textbf{Depth Prediction on Make3D test set.}}}
\label{fig:make3d}
\end{center}
\end{figure*}

\begin{figure*}[ht!]

\begin{center}

\begin{subfigure}{0.96\textwidth}
  \begin{center}
  \includegraphics[scale=0.95]{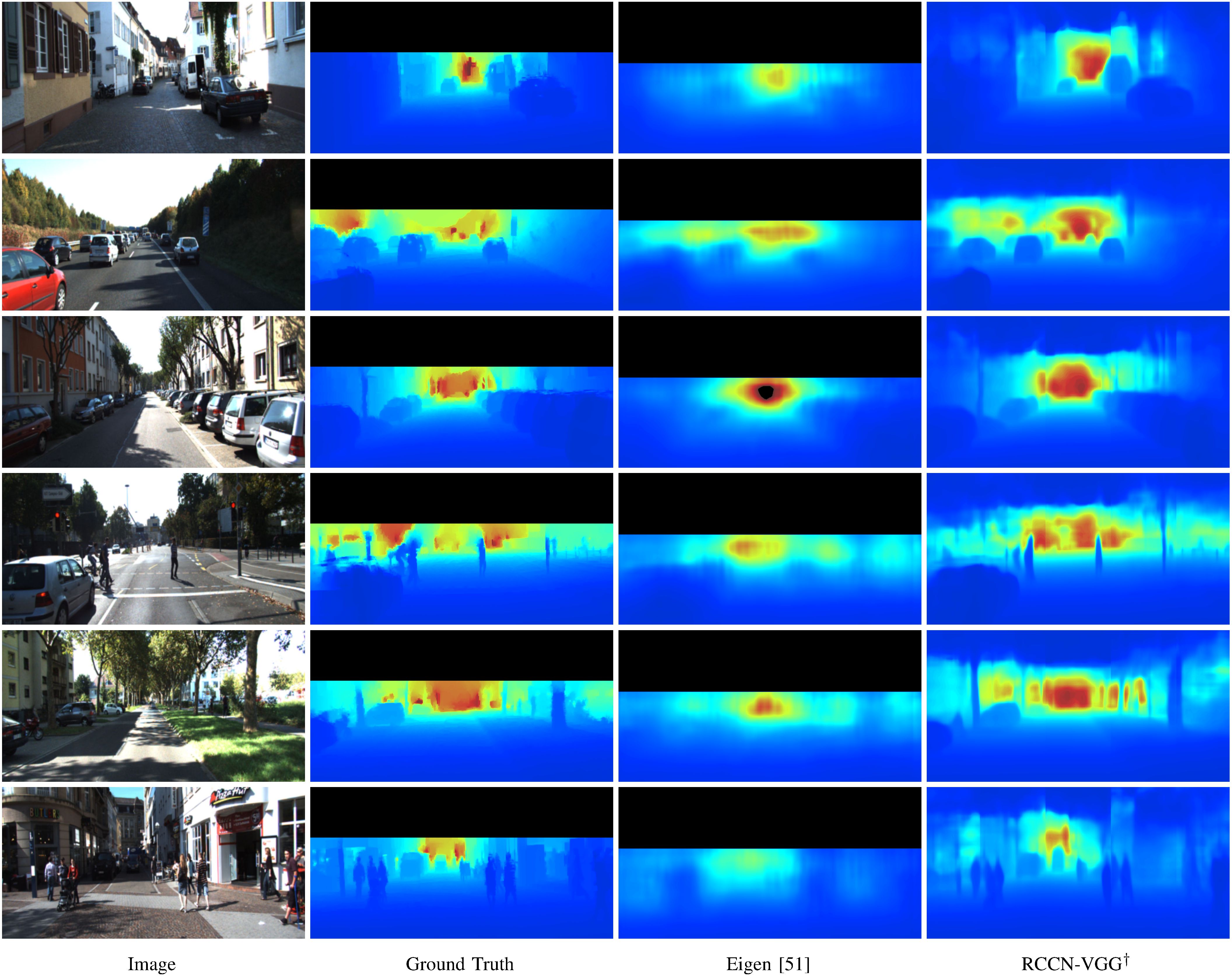}
  \end{center}
\end{subfigure}%

\caption{\small{\textbf{Depth Prediction on KITTI test set.} The depth values in the black parts are not provided by KITTI.}}
\label{fig:kitti}
\end{center}
\end{figure*}

\subsection{KITTI}
\label{exp:kitti}
The KITTI dataset \cite{Geiger2013IJRR} contains some outdoor scenes captured by cameras and depth sensors in a driving car. All the 61 scenes from the ``city", ``residential", and ``road" categories consist of our training/test sets.  We test on 697 images from 29 scenes split by \cite{eigen2014depth}, and train on 23,486 images from the remaining 32 scenes. All the images are resize to a resolution of $385 \times 769$ from $375 \times 1241$.
We train our model on a random crop of size $385 \times 385$. At the test time, we split each image to 3 overlapping windows and obtain the predicted depth values in overlapped regions by averaging the 3 predictions. The evaluation metrics are computed on a pre-defined center cropping by Eigen \etal \cite{eigen2014depth} in the original resolution. Note that, since the ground truth depth are provided for only about 15\% of points within the bottom part of the image, some depth targets in the bottom parts are filled using the colorization routine in the NYU Depth development kit \cite{levin2004colorization} for our training images following \cite{eigen2014depth}.

Besides those representative quantitative results shown in Fig.~\ref{fig:kitti}, we summarize the quantitative evaluation in Tab.~\ref{tab:kitti}, which demonstrates that the proposed approach significantly outperforms those previous methods in all those considered error metrics.


To further exploit the compromise principle and demonstrate the effectiveness of the proposed RCCN, we learn some related network variants by directly regressing continuous depth, directly estimating discrete depth via classification, jointly modeling continuous depth via our network architecture, and learning a classification-classification cascade network. From the quantitative results shown in Tab.~\ref{tab:varient} and Fig.~\ref{fig:loss}, we can conclude that: 1) when treating depth estimation as a classification problem instead of regression, the network can converge to a better local solution on average; 2) image-receptive-field-of-view understanding, as well as local high-level convolutional features indeed help deep networks better learn the depth distribution of a scene; 3) the compromise between spatial and depth resolutions simplifies network training; 4) The performances of RRCN and of RCCN deteriorate once removing the low-spatial resolution branch $D_{0}$ from them (leading to R, C, respectively), demonstrating that $D_{0}$ benefits $D_{1}$; 5) The comparison between $D_{0}$ and RCCN-$D_{0}$ shows that $D_{1}$ also benefits $D_{0}$;  and 6) From the observation that RCCN $>$ CCCN $>$ RRCN, It can be seen that high spatial resolution together with high depth resolution leads to worse results.

\subsection{NYU Depth V2}

The NYU Depth V2  \cite{Silberman:ECCV12} contains 464 indoor video scenes taken with a Microsoft Kinect camera.  We randomly sample half of the 120K images from the raw dataset according to the official split training scenes as our training sets, and test on the 694-image test set. We train our model on a randomly crop of size $240 \times 320$.
On top of those qualitative results in Fig. \ref{fig:nyu}, we report in Tab. \ref{tab:nyuv2} the quantitative results via several common metrics used in previous works \cite{eigen2015predicting, liu2015deep}. 
The predictions from our model yields comparable or state-of-the-art results comparison with previous works. 
Specifically, the estimated coarse continuous depth outperforms the ``Coarse+Fine" prediction of Eigen \etal \cite{eigen2014depth}. The predicted high spatial resolution discrete depth in particular obtains an impressive improvement, demonstrating that both discrete depth and the proposed RCCN framework are effective.
\begin{figure*}[ht!]

\begin{center}

\begin{subfigure}{0.96\textwidth}
\begin{center}
  \includegraphics[scale=0.96]{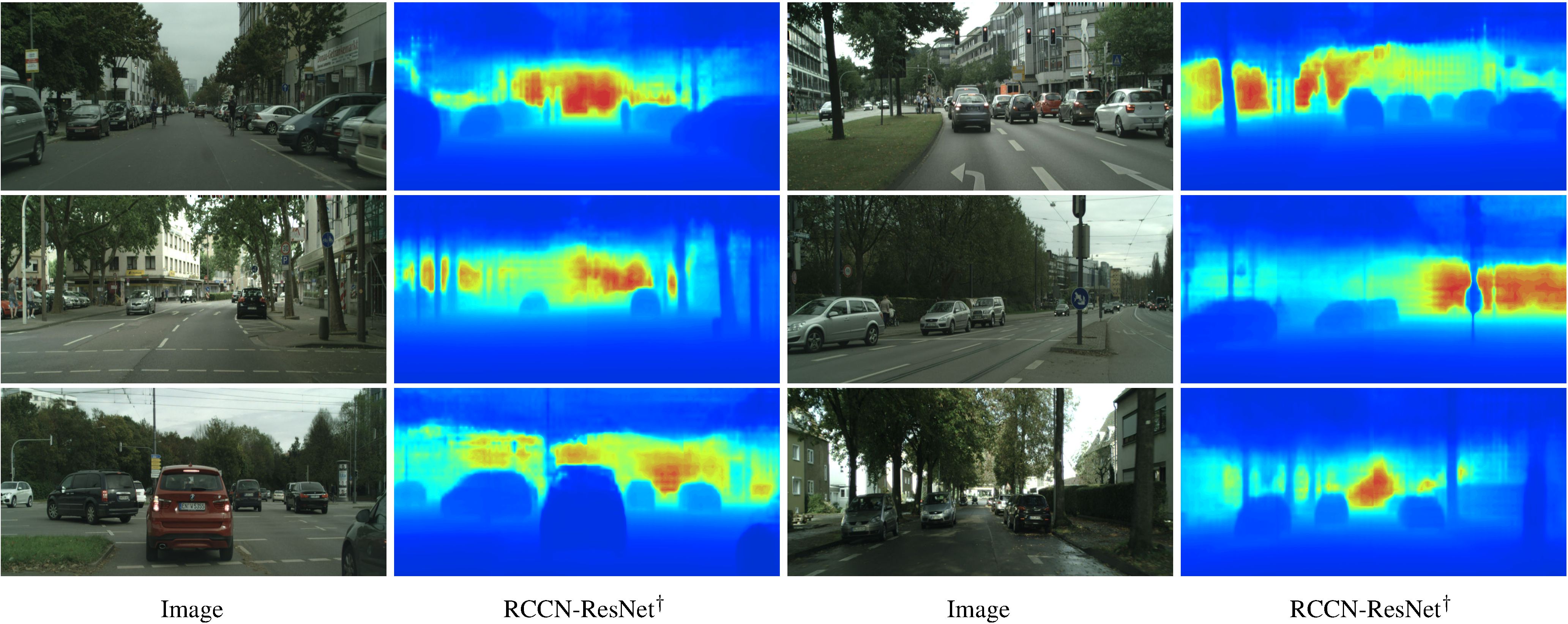}
  \end{center}
\end{subfigure}%
\caption{\small{\textbf{Generalization to Cityscapes.} The model is trained only on KITTI, and test on Cityscapes.}}
\label{fig:city}
\end{center}
\end{figure*}

\subsection{Make3D}

The Make3D dataset \cite{saxena2006learning, saxena2009make3d} contains 534 outdoor images, 400 for training, and 134 for testing, with the resolution of $2272 \times 1704$. Our model is trained on a random crop of size $480 \times 320$. In the test phase, we split the test images into some $480 \times 320$ sub-images, and use max pooling on the overlapping regions to obtain the final predictions. As shown in Tab.~\ref{tab:make3d}, we report \emph{C1} and \emph{C2} error on this dataset \cite{karsch2014depth}. We achieve state-of-the-art performance in all error metrics.

\subsection{Generalization to Cityscapes}
We also demonstrate the generalization ability of our model. Specifically, we test our model trained only on KITTI via images provided by Cityscapes \cite{Cordts2016Cityscapes}, which is also a large benchmark for auto-driving. As shown in Fig.~\ref{fig:city}, our KITTI model can capture the general scene layout and objects such as cars, trees and pedestrians much well in images from Cityscapes.



\section{Conclusion}
\label{sec:conclusion}
In this paper, we have presented a deep CNN architecture for~\ProblemAbbr. Based on the fact that training a network to estimate a high spatial resolution continuous depth map is difficult, we hypothesize to design network architectures according to the compromise principle that training a network to estimate a depth map with reduced spatial resolution or depth resolution is easier.  According to the compromise principle, we propose a regression-classification cascaded network to jointly model continuous depth and discrete depths in two branches. The proposed approach is validated on three widely-used and challenging datasets, where it achieves competitive or state-of-the-art results.  Moreover, specific experiments have been done and the obtained results demonstrate that our network is superior to its variants, which also validates the design of our approach to some extent. We will continue to investigate new methodologies to reduce the depth resolution and extend our framework to other challenging dense prediction problems.


%





\ifCLASSOPTIONcaptionsoff
  \newpage
\fi



\bibliographystyle{IEEEtran}
\bibliography{depthprediction}

\end{document}